\documentclass[final]{cvpr}

\usepackage{times}
\usepackage{epsfig}
\usepackage{graphicx}
\usepackage{amsmath}
\usepackage{amssymb}
\usepackage[font=small,skip=0pt]{caption}
\usepackage{lipsum} % added for dummy texts
\usepackage{booktabs}
\usepackage{enumitem}
\usepackage[page,header]{appendix}
\usepackage{multibib}

\newcommand{\devc}{Zhang {\it et al.} \cite{devc}}

\usepackage[pagebackref=true,breaklinks=true,colorlinks,bookmarks=false]{hyperref}

\def\cvprPaperID{0000} % *** Enter the CVPR Paper ID here
\def\confYear{CVPR 2021}

\begin{document}

%%%%%%%%% TITLE
\title{Reference-Based Video Colorization with Spatiotemporal Correspondence}

\author{Naofumi Akimoto \quad Akio Hayakawa \quad Andrew Shin \quad Takuya Narihira\\
Sony Corporation\\
Tokyo, Japan\\
%{\tt\small firstauthor@i1.org}
% For a paper whose authors are all at the same institution,
% omit the following lines up until the closing ``}''.
% Additional authors and addresses can be added with ``\and'',
% just like the second author.
% To save space, use either the email address or home page, not both
}

\twocolumn[{%
\renewcommand\twocolumn[1][]{#1}%
\maketitle 
\begin{center}
%\centering
\vspace*{-20pt}
\includegraphics[width=\textwidth]{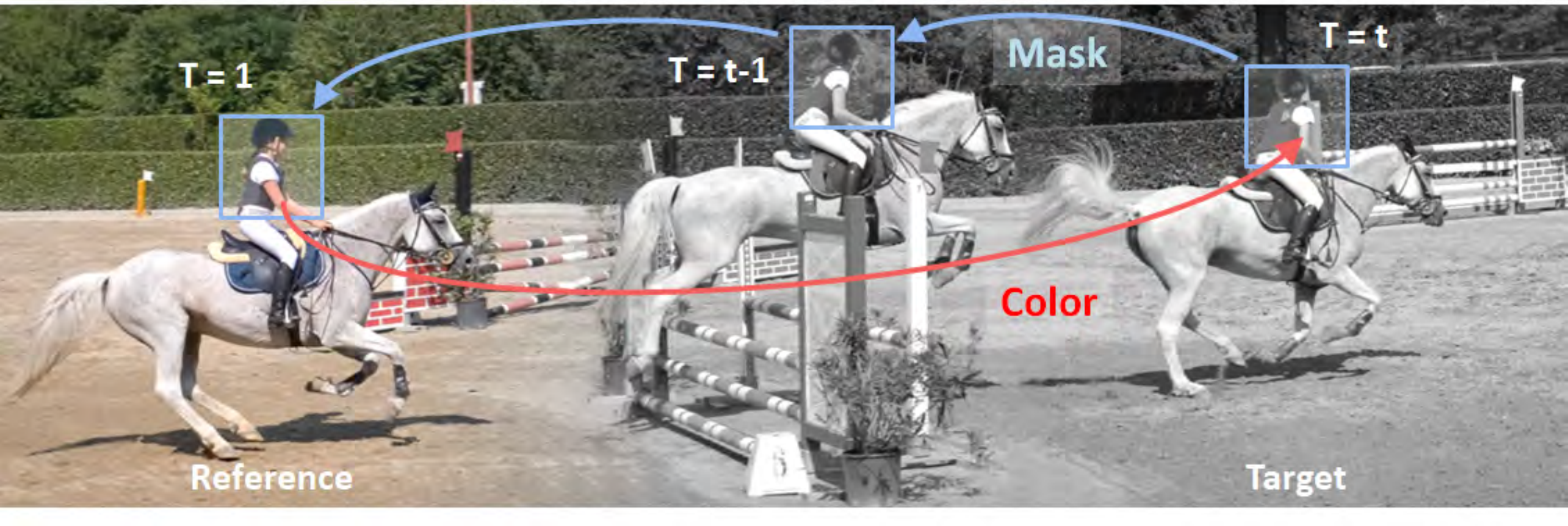}
\captionof{figure}{
    We propose a novel reference-based video colorization framework with spatiotemporal correspondence. We propagate a mask as correspondence in time, and then warp colors with spatial correspondence only from the regions on the reference frame restricted by the temporal correspondence. Our framework realizes faithful and long-term colorization.
    }
\label{fig:teaser}
\end{center}%
}]

%\maketitle

%%%%%%%%% ABSTRACT
\begin{abstract}
   We propose a novel reference-based video colorization framework with spatiotemporal correspondence.  Reference-based methods colorize grayscale frames referencing a user input color frame. Existing methods suffer from the color leakage between objects and the emergence of average colors, derived from non-local semantic correspondence in space. To address this issue, we warp colors only from the regions on the reference frame restricted by correspondence in time. We propagate masks as temporal correspondences, using two complementary tracking approaches: off-the-shelf instance tracking for high performance segmentation, and newly proposed dense tracking to track various types of objects. By restricting temporally-related regions for referencing colors, our approach propagates faithful colors throughout the video. Experiments demonstrate that our method outperforms state-of-the-art methods quantitatively and qualitatively.
\end{abstract}

%%%%%%%%% BODY TEXT
% ---------------------------------------------
\section{Introduction}
Colorizing black-and-white videos has increasingly gained attention, since it can revitalize a massive amount of outdated contents into modern standards. %, e.g., re-commercialize old contents in movie industry. % along with historical value. 
Previously, production of high-quality contents has required manual colorization, resulting in prohibitively high costs in terms of time and labor. Researchers have thus endeavored to come up with a scheme to automate the colorization process, in fully automatic \cite{Lei_2019_CVPR} or semi-automatic ways \cite{deepremaster, devc}. 
%Regardless of the specific colorization model, the primary issue has been the quality of the output as well as usability.
Yet, video colorization remains a highly challenging task. On top of the challenging aspects of image colorization, such as its ill-posed nature with a large number of potentially correct results, video colorization imposes additional dimensions of challenge, namely maintaining temporal coherence, computational cost, and user controllability. 
%In particular, user controllability becomes a critical factor when it comes to practical usage. %As such, user-guided approach \cite{deepremaster, devc} can come in handy when dealing with the task, since it designates specifically which color is to appear. It is also easier to preserve temporal coherence than fully automatic approaches.

%In conventional approaches (cite), a commonly used technique was to propagate the flow obtained from the color of strokes, but it is difficult to obtain an accurate flow from videos with relatively longer duration. 
%Another approach is a reference-based approach, in which a key frame is colorized, and the color is reflected throughout the remaining frames. For example, propagation-based methods (cite) predict the movement of pixels from neighboring frames, and propagate the color from the previous frame in order. However, this model can easily accumulate mispredictions as it propagates further, making it difficult to reflect the user-designated color to distant frames. Such incorrect propagation is the most challenging problem in propagation-based approach.

%In order to deal with such problem, 
Recent attention-based approaches \cite{deepremaster, devc} proposed to compute correspondence between the target frame and reference frame using semantic features, and colorize the target frame by attending to the color from reference frame. This approach can alleviate the issue of accumulating mispredictions over the frames, which is prevalent in propagation-based approaches \cite{jampani_vpn_2017, liu2018switchable}.
%On the other hand, it disregards the aspect of maintaining temporal correspondence over continuous frames. 
However, as shown in Fig. \ref{fig:quality_single}, its non-local spatial correspondence frequently results in color leakage between instances, or the objects colorized with average colors from multiple objects.

To address this issue, we instead propose a reference-based video colorization with spatiotemporal correspondence. The key assumption of our work is that, by assigning color to objects that exist in both reference and target frames with finer correspondence, we can enhance video colorization, as illustrated in Fig.~\ref{fig:teaser}. 
However, if attention-based model is augmented in an incremental manner by warping the color from both reference frame and previous frame, it inevitably results in placing higher priority for previous frame, since it tends to resemble the target frame more. As a result, it is likely to be degraded to propagation-based model. We thus need another approach to account for the \textit{continuity} of regions over temporal axis, while avoiding the errors from accumulated color propagation.

\begin{figure}
%\begin{center}
\centering
\includegraphics[width=0.8\linewidth]{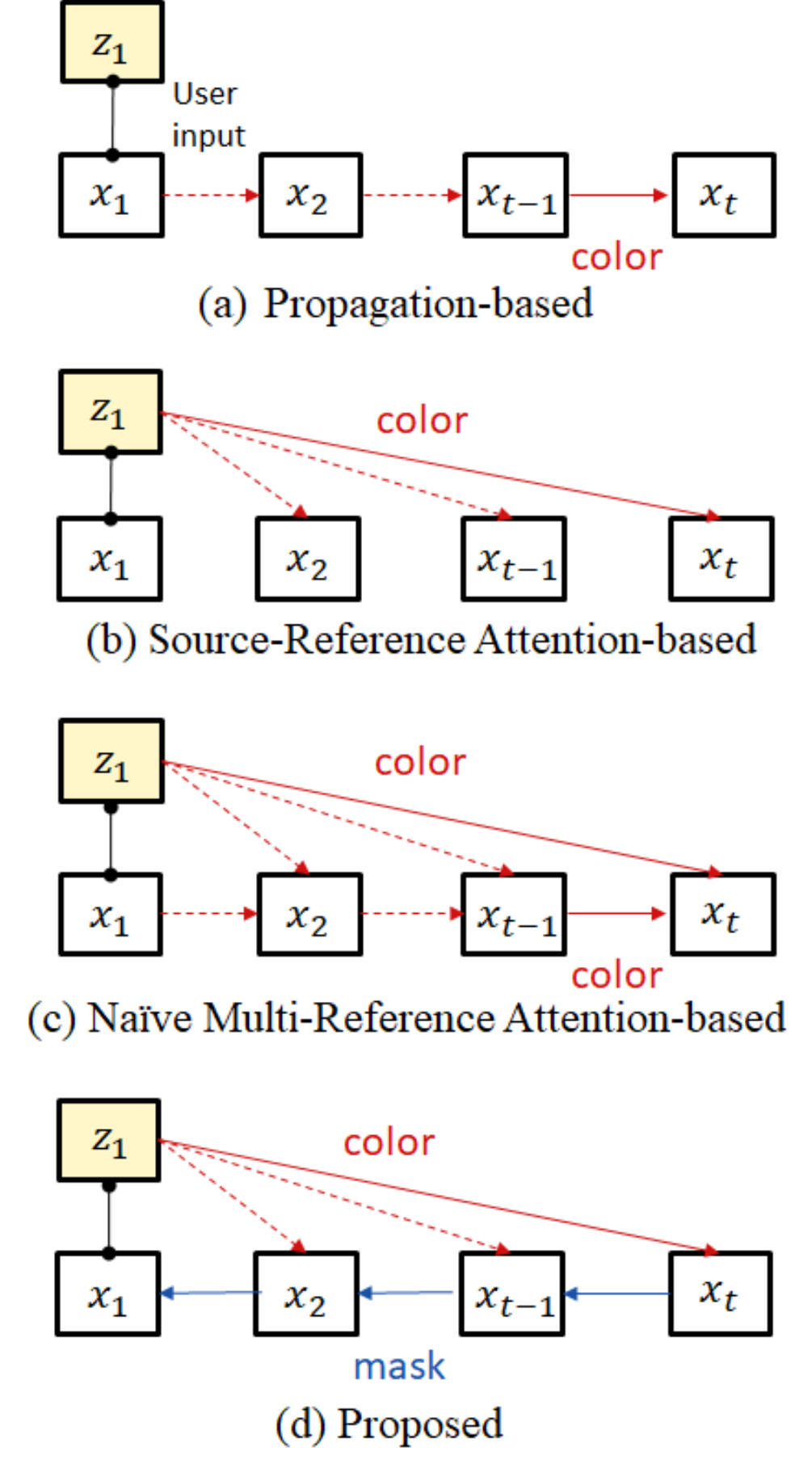}

   \caption{Comparison among four types of approaches when processing an input video $x$ with reference frame $z$. The figures illustrate a user colorizing $x_1$ with a reference frame $z_1$. (a) and (c) suffer accumulation error, because they colorize $x_t$ with colored $x_{t-1}$. See Fig. \ref{fig:temporal_corr} for a visualization of (c). (d) warp colors considering temporal correspondence between $x_t$ and $z_1$, unlike (b).}
\label{fig:approach}
%\end{center}
\end{figure}

Our key ideas to tackle this technical issue are as following: 1) propagate masks to represent the correspondence over time axis instead of color, and 2) color is determined within the region restricted by the mask in the reference frame. Fig.~\ref{fig:approach} illustrates the differences between our model and others.
Our framework consists of three stages, namely tracking, warping, and refinement. 
Tracking is performed in order to obtain the temporal correspondence between reference frame and target frame. We can specify which color to warp more accurately, by specifying the region in reference frame that is temporally correlated to pixels to colorize on target frame. 
For warping color, we obtain semantic features by feeding reference and target frames to the network, and compute the spatial correspondence based on their similarity. Furthermore, by warping the color solely from the region specified by tracking, we can warp the color based on spatiotemporal correspondence. In contrast to propagation-based models in which the succeeding frames may be colorized with incorrect color due to accumulation of prediction error over the frames, our model can perform colorization with vivid color specified by reference frame via propagating masks rather than colors.
Finally, refinement stage employs encoder-decoder network to recover the mistakes from warping colors and enhance the temporal coherence throughout neighboring frames.

%%%%% Figure %%%%%
%%%%%

Moreover, we propose to incorporate two types of tracking as a way to obtain correspondence over temporal axis, which restricts the region to warp color from. We first employ 1) instance tracking in order to prevent incorrectly blending the colors from distinct objects. To the best of our knowledge, our model is the first video colorization work that is instance-aware. By using instance tracking, we can accurately differentiate between instances. However, instance tracking alone cannot track objects of undefined classes, or track at granularity smaller than the instance. In order to resolve such issues, 2) we propose a mechanism to track at pixel-level. Inspired by dense tracking \cite{tracking_emerges}, we exploit correspondence learned from colorization for tracking without having to further train the model. %Note that our model is designed in a novel way to enhance performance on colorization task specifically. For example, there may be cases where object to track is not clear. 
By specifying temporally correlated regions for each pixel, we can avoid the problem of blurry colors resulting from warping color from the entire frame.

Through the approaches above, our proposed model is capable of high-quality colorization for videos of long duration with complex scenes, frequently involving change in background or rapid movement of objects. Its faithful reflection of reference also leads to reduced amount of manual work and enhanced user controllability. Experiment results clearly demonstrate that our model outperforms state-of-the-art-models both quantitatively and qualitatively. In particular, our spatiotemporal correspondence turns out to be effective when dealing with complex scenes that attention-based approach finds difficult. %We also thoroughly examine and discuss our spatio-temporal correspondence, which shows the effectiveness of incorporating tracking at instance and pixel-level. %We highly anticipate our new framework to accelerate the application of video colorization.

Our contributions can be summarized as following: 
\begin{itemize}[noitemsep,topsep=0pt]
\itemsep0em 
\item We propose a novel reference-based video colorization framework with spatiotemporal correspondence. We propagate a mask as correspondence in time, and warp colors with spatial correspondence from the regions restricted by the temporal correspondence.
\item We propose a novel dense tracking specialized for video colorization. With no additional training, it faithfully reflects the color from user reference.
\item Our proposed method outperforms state-of-the-art methods and improves user-controllability, while alleviating color leakage or color being averaged out.
\end{itemize}
% ---------------------------------------------

% ---------------------------------------------
\section{Related Work}
\textbf{Image Colorization.} Early image colorization was mostly performed with user-guided signatures and sample references \cite{Dani04colorizationusing, HuangTCWW05, EGWR:EGSR07:309-320, qu-2006-manga}. These methods propagate local color guidance to similar regions using low-level features.
They frequently relied on deep-learning in fully automatic manner to learn the semantic-color relationships from a large dataset \cite{ IizukaSIGGRAPH2016, pix2pix2017, zhang2016colorful}, or to propagate the user-guided scribbles in the spatial direction based on the semantic feature \cite{zhang2017real}, or warp the color to a corresponding position based on exemplar color sample \cite{he2018deep}.
Recently, as a new perspective, an instance-aware approach has been proposed for image colorization \cite{instance-aware}. Learning instance-color relationships improves the performance of fully automatic image colorization. We extend the instance-aware approach for video colorization.

\textbf{Video Colorization.} Video colorization is far more difficult than image colorization and is still an open challenge.
An intuitive approach is to perform post processing \cite{bonneel2015blind, Lai-ECCV-2018} to impose temporal coherence to flickers of per-frame colorization, but the colors are likely to be washed out. 
%The propagation of color scribbles across frames using optical flow may encounter cases in which the scribbles are not suitable for other frames.The method of colorizing first frame and propagating following frames suffers from accumulation of errors, and consequently, the number of frames that can be propagated is limited.
The propagation of color scribbles \cite{levin2004colorization, yatziv2006fast} or a colored first frame \cite{jampani_vpn_2017, liu2018switchable, tracking_emerges} across frames using estimated flow may suffers from accumulation of errors, and consequently, the number of frames that can be propagated is limited.

Recently, source-reference attention-based methods \cite{deepremaster, devc} has been proposed. Accumulation of errors is reduced by using a reference image for coloring each frame instead of using the last colored frame. However, there exists no direct handling of temporal information for correspondence. These methods warp colors solely based on the relationship to a reference frame, and then consider the relationship between target frames for temporal coherence. Due to non-local attention, colors are frequently mixed even between objects that should be distinguished.
Conversely, our method warps a color from a part of a reference frame by limited correspondence based on temporal information of the target frame. 
This makes it possible to determine a color only from a similar region in the nearest neighbor in space and time, resulting in a colorization without a color leakage between instances and a faithful colorization for the reference image.

%% dense tracking 
\textbf{Dense Tracking.} Dense tracking is a task that re-localizes objects in the preceding and succeeding frames with pixel-level segmentation masks. Recent self-supervised learning methods trained with video colorization \cite{jabri2020space, lai2020mast, li2019joint, tracking_emerges} or cycle consistency \cite{wang2019unsupervised, CVPR2019_CycleTime} propagate a given segmentation mask to subsequent frames based on the correspondences. In video colorization task, however, we cannot use methods that require pre-processed object's masks provided by user as input \cite{caelles2017one, luiten2018premvos, maninis2018video}, or methods that use color information \cite{jabri2020space, lu2020learning}.
To improve colorization at pixel-level, we need masks for each pixel of the target frame separately, rather than tracking several representative object masks.
Our newly proposed dense tracking method is designed to be applicable to such requirements of colorization task, inspired by the method \cite{tracking_emerges}, as we describe in Sec.~\ref{sec:dense}.

\begin{figure*}
%\begin{center}
\centering
\includegraphics[width=\linewidth]{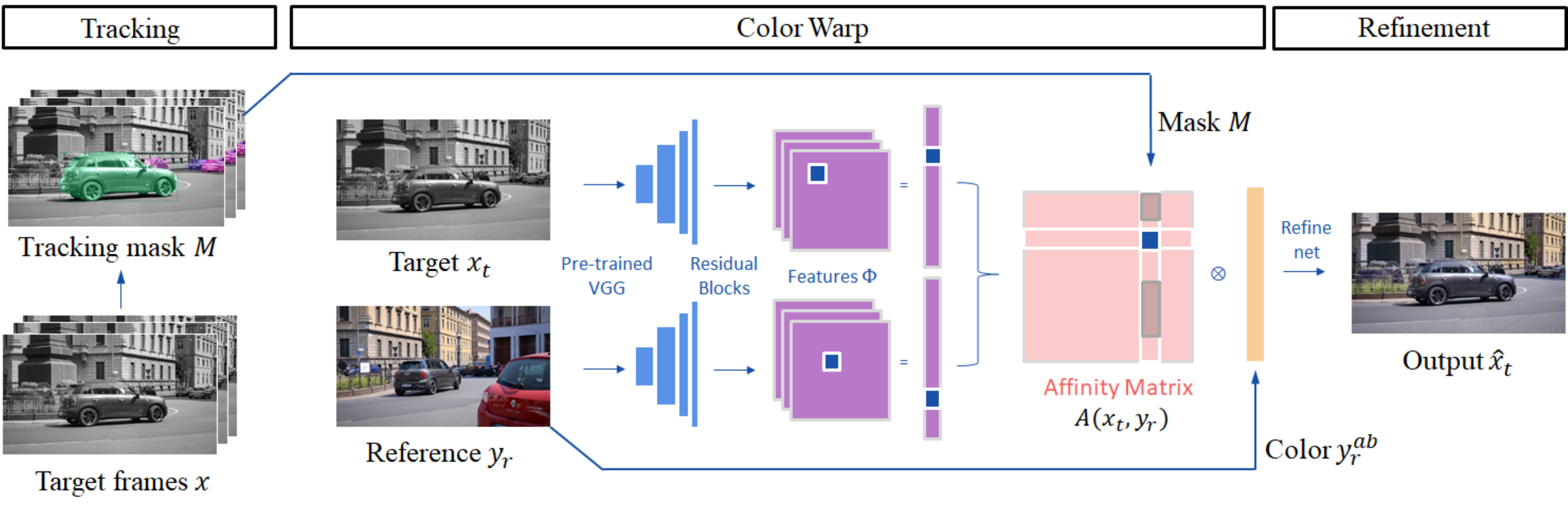}
   \caption{Pipeline overview. Our framework consists of three stages: tracking stage for obtaining a mask as a temporal correspondence between target frames (Sec. \ref{sec:instance} and Sec. \ref{sec:dense}), color warping stage for coloring a target frame with the tracking mask (Sec. \ref{sec:spatially-restricted}, and refinement stage for temporal coherence.}
\label{fig:model}
%\end{center}
\end{figure*}

% ---------------------------------------------
\section{Method}
% ---------------------------------------------
\subsection{Overview}\label{method:overview}
We propose a reference-based video colorization framework which consists of three stages; a tracking stage, a spatially restricted color warping stage, and a color refinement stage. Fig.~\ref{fig:model} shows the overview of our framework.
Inputs for the whole system are grey-scale frames $x^l = \{x^{l}_t\}_{t=1}^{t=N}$ where $x^{l}_t \in \mathbb{R}^{H \times W \times 1}$, and some colored reference frames picked from the video are $y^{ab} = \{y^{ab}_r\}_{r = 1}^{N_{\mathrm{ref}}}$ where $y^{ab}_r \in \mathbb{R}^{H \times W \times 2}$, and the outputs are the chrominance {\it ab} for each frame $\hat{x}^{ab} = \{\hat{x}^{ab}_t\}_{t=1}^{t=N}$ where $\hat{x}^{ab}_t \in \mathbb{R}^{H \times W \times 1}$.

To obtain a mask for restricting regions to calculate correspondence, we execute instance tracking (Sec. \ref{sec:instance}) and dense tracking (Sec. \ref{sec:dense}). 
We warp colors based on an affinity matrix $A \in [0, 1]^{HW \times HW}$ between semantic features of a target frame $x^l_t$ and a reference frame $y^{l}_r$.
To colorize the target frame with colors faithful for the reference frame, we calculate a semantic correspondence only among regions which have temporal relations, instead of non-local semantic correspondence (Sec. \ref{sec:spatially-restricted}). To do so, we use the tracking masks for each reference frame $M = \{M_r\}_{r = 1}^{N_{\mathrm{ref}}}$ where $M_r \in \{0, 1\}^{HW \times HW}$, which captures temporal information among the target frames. Furthermore, we extend this mechanism to handle multiple reference images. Namely, we calculate warped colors $W^{ab}_t \in \mathbb{R}^{H\times W \times 2}$ for target frame $x_t$ by warping function $F$ as follows:
\begin{equation}
    \label{eqn:warping function}
    W^{ab}_t = F(x^l_t, y^{ab}_1, M_1, \dots, y^{ab}_{N_{\mathrm{ref}}}, M_{N_{\mathrm{ref}}})
\end{equation}
Note that $M_r$ here represents tracking result of $y_r$ propagated from t-th frame $x_t$. For all frames $\{x_t\}_{t=1}^{N}$, we calculate different masks from target frames to each reference frame in the same way. For simplicity, we hereinafter omit $x_t$ and $y_r$ and refer to $M_r$ as a mask between a target frame $x_t$ and a reference frame $y_r$.

In color refinement stage, we use fully-convolutional u-net proposed in Zhang {\it et al.} \cite{zhang2016colorful} . This network refines the warp result to have temporal coherence with the consecutive frame and improves spatial consistency. The inputs for the refinement network are the colorized last frame $\boldsymbol{\hat{x}}_{t-1}$, the warp result of the frame to be colorized $W^{ab}_t$, and the warp confidence map $S_t \in [0, 1]^{H\times W}$ that is max values of the affinity matrix for the reference dimension. This refinement function $G$ can be expressed as:
\begin{equation}
    \hat{x}_t^{ab} = G(x_{t-1}^l, x_t^l, \hat{x}_{t-1}^{ab},  W^{ab}_t, S_t) 
\end{equation}

% ---------------------------------------------
\subsection{Spatially restricted Correspondence} \label{sec:spatially-restricted}
In this section, we present the calculation of semantic correspondence at first, and the calculation of the correspondence restricted by the tracking mask with temporal information, and extend the mechanism to handle multiple references.

To obtain the semantic correspondence between target frame $x_t$ and reference frame $y_r$, we transfer the target frame and the reference frame to VGG network \cite{vgg} features, and concatenate features from multiple layers. Feeding them into further residual blocks, we extract features $\Phi(x_t), \Phi(y_r) \in \mathbb{R}^{HW \times L}$, respectively.  Note that we use only luminance {\it l} for both $x_t$ and $y_r$ to calculate $\Phi(x_t), \Phi(y_r)$. Using $\Phi(x_t)$ and $\Phi(y_r)$, we calculate the affinity matrix $A$ of the target frame and the reference frame from correlation matrix $C \in \mathbb{R}^{HW \times HW}$ as follows:
\begin{align}
    A(x_t, y_r) =&\  \underset{j}{\text{softmax}}(C(x_t, y_r)) \\
    C_{i,j}(x_t, y_r) =&\  \frac{(\Phi_{i}(x_t) - \mu_{\Phi_{x_t}})^{\mathrm{T}} (\Phi_{j}(y_r) - \mu_{\Phi_{y_r}})}{\|\Phi_{i}(x_t) - \mu_{\Phi_{x_t}}\|_2\ \|\Phi_{j}(y_r) - \mu_{\Phi_{y_r}}\|_2}
\end{align}
where, $\mu_{\Phi_{x_t}}$ and $\mu_{\Phi_{y_r}}$ represent mean feature vectors. 
Based on this affinity matrix, we calculate warped colors from reference frame to target frame by the product of the affinity matrix and the \textit{ab} vector of the reference frame. Thus, in the setting with a single reference frame, equation \ref{eqn:warping function} can be represented as:
\begin{equation}
W^{ab}_{t}(i) = \sum_{j} A_{i,j}(x_t, y_r)\ y^{ab}_{r}(j)
\end{equation}
where $W^{ab}_{t}(i) \in \mathbb{R}^{2}$ is a chrominance {\it ab} of the $i$-th position at $t$-th frame and $y_r^{ab}(j) \in \mathbb{R}^{2}$ is that of the $j$-th position at a reference frame $y^{ab}_r$. 

This algorithm relates the target frame and the reference frame to each other in non-local regions, without considering their temporal relation. However, our objective is to determine the colors only from the regions that are estimated to be relevant based on the time information. To do so, we use a tracking mask based on the time series relationship for this affinity matrix to limit the candidates. Specifically, we replace $A$ in equation \ref{eqn:warp_single} with a restricted affinity matrix $\tilde{A } \in [0, 1]^{HW \times HW}$ that is:
\begin{align}
\label{eqn:warp_single}
W^{ab}_{t}(i) = \sum_{j} \frac{\tilde{A}_{i,j}(x_t, y_r, M_r)}{\sum_{i'}\tilde{A}_{i',j}(x_t, y_r, M_r)}\ 
y^{ab}_{r}(j) \\
\tilde{A }_{i,j}(x_t, y_r, M_r) = 
\begin{cases}  
A_{i,j}(x_t, y_r)& \text{if } M_{r,i,j} = 1 \\ 
0, & \text{otherwise}  
\end{cases} 
\end{align}
This suggests that we require the masks $M_r$ on the reference frame to correspond to each target pixel. As a way to prepare tracking masks containing temporal relations, we propose two approaches: using masks from instance tracking (Sec. \ref{sec:instance}) and using masks from a dense tracking (Sec. \ref{sec:dense}). We describe masks obtained by these two approaches as $M^I$ and $M^D$ in the following sections.

Considering a practical application, the method using a single reference as in \cite{devc} is likely insufficient because there are many colors that cannot be specified when working with long videos in which many new colors and objects appear. To address this issue, we extend the model to deal with multiple references. In this case, the equation \ref{eqn:warp_single} can be extended by multiple references $y$ and masks $M$ as follows:
\begin{equation}
W_t^{ab}(i) = \sum_{r,j} \frac{\tilde{A}^{m}_{r,i,j}(x_t, y,M)}{\sum_{i'}\tilde{A}^{m}_{r,i',j}(x_t, y,M)}\
y_{r}^{ab}(j)
\end{equation}
where $\tilde{A}^m, A^m \in [0, 1]^{N_{\mathrm{ref}} \times HW \times HW}$ is a restricted and non-restricted stacked affinity matrix and $C^m \in \mathbb{R}^{N_{\mathrm{ref}} \times HW \times HW}$ is a stacked correlation matrix such as:
\begin{align}
    \tilde{A}^m_{r,i,j}(x_t, y, M) =& 
    \begin{cases}
        A^m_{r,i,j}(x_t, y)& \text{if } M_{r,i,j} = 1 \\ 
        0, & \text{otherwise} \\
    \end{cases} \\
    A^m(x_t, y) =&\  \underset{r, j}{\mathrm{softmax}}(C^m(x_t, y)) \\
    C^m(x_t, y) =&\  \mathrm{stack}(C(x_t, y_1), \dots, C(x_t, y_{N_{\mathrm{ref}}}))
\end{align}
%While DeepRemaster \cite{deepremaster} fuses the colors to features once, our multi-reference mechanism directly warps colors from multiple references.
%Thus, our method successfully produces more faithful and vivid output colors.
% Our multi-reference mechanism directly warps colors from the reference images, so it produces faithful and vivid output colors than the DeepRemaster \cite{deepremaster}, which fuses the color to features once.

% ---------------------------------------------
\subsection{Instance Tracking}\label{sec:instance}
To solve the problem of warping color between instances,
we use off-the-shelf pre-trained instance segmentation \cite{maskrcnn} to calculate an instance mask for each frame, and then use IoU (Intersection over Union) of each mask between adjacent frames to identify the same instance.
First, using this tracking function $T$, we obtain instance tracking masks $\{I'_t(k)\}_{t=1}^{N} \in [0, 1]^{HW}$ as vectors for each object $k \in \{1, 2, \dots, N_{\mathrm{obj}}\}$:
\begin{equation}
I'_{1}(k),\dots, I'_{N}(k)  = T(I_{1}(k), \dots , I_{N}(k))
\end{equation}
where $\{I_t(k)\} \in [0, 1]^{HW}$ is an instance mask of object $k$ in t-th frame.
Let $\phi_{i}(x_t) \in \{1, 2, \dots, N_{\mathrm{obj}}\}$ be a function to get a label id at $i$-th position of target frame $x_t$. Finally, we calculate an instance tracking mask $M^I \in [0, 1]^{HW \times HW}$ for a reference frame $y_r$ to limit affinity between $x_t$ and $y_r$ as follows:
\begin{equation}
M^{I}_{r,i,j} = 
\begin{cases}  
1& \text{if } I'_{r,j}(\phi_{i}(x_t)) \geq \mathrm{threshold} \\ 
0& \text{otherwise}  
\end{cases}
\end{equation}

%%%%% Figure %%%%%
\begin{figure}
\begin{center}
\includegraphics[width=\linewidth]{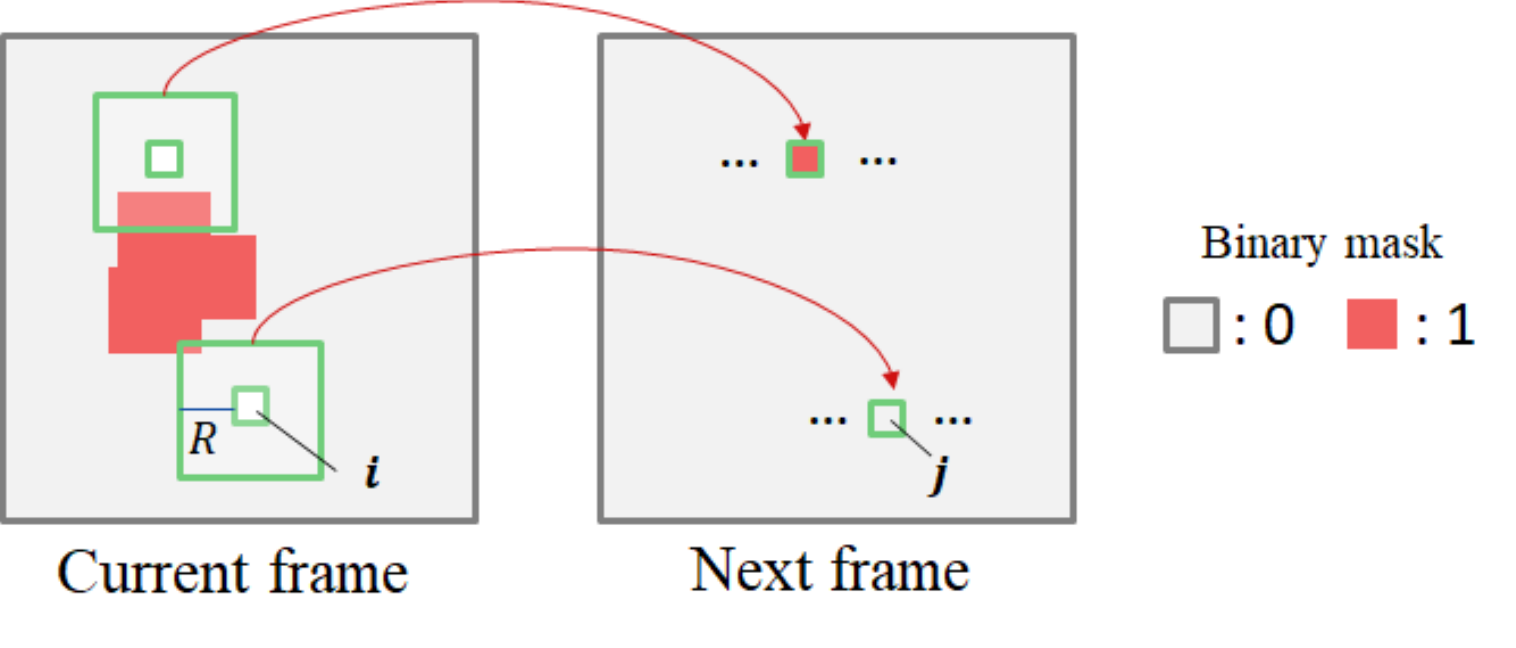}
\end{center}
   \caption{Mask propagation by dense tracking. Our method propagates a mask to the next frame when the similarity between a pixel on the next frame and the neighborhood on the current frame exceeds a threshold.}
\label{fig:algo_dense}
\end{figure}
%%%%%
% ---------------------------------------------
\subsection{Dense Tracking}\label{sec:dense}
Because instance tracking can only handle instances of classes that are already defined, it is not possible to capture temporal relationships of regions of non-defined classes, or areas other than the instance.
To compensate for this limitation and make the color of the reference frame faithful across the image, we introduce dense tracking.
Previous studies of self-supervised dense tracking have shown that trackers can be learned by training reference-based colorization \cite{tracking_emerges}. Since we already have a correspondence mechanism for video colorization, we can use dense tracking without additional training.

The difference in setting from \cite{tracking_emerges} is that it is necessary to individually track each pixel of the frame to be colored, instead of tracking a set of pixels. In \cite{tracking_emerges}, a set of pixels is specified as an object mask for the first frame to track. Conversely, in our task, the individual pixels of the target to be colored are the target of tracking. We thus need to propose a new dense tracking algorithm since the mask would disappear immediately when we directly use the existing method for our task.
% when the existing method is directly used for our task.

%%%%%%%%%% Table
\begin{table*}
\begin{center}
\begin{tabular}{lccccccccc}
\toprule
&\multicolumn{3}{c}{Single Reference}&
\multicolumn{6}{c}{Multiple References}\\
\midrule
&\multicolumn{3}{c}{10th frame}&
\multicolumn{3}{c}{1st \& Nth frames}  & \multicolumn{3}{c}{10th \& N-10th frames} \\
Method & Full & Inner & Outer & Full & Inner & Outer & Full & Inner & Outer \\
\midrule
DeepRemaster \cite{deepremaster}& 25.53 & 24.44 & 25.70 & 25.38 & 24.18 & 25.60 & 25.53 & 24.40 & 25.73 \\
\devc{} & 28.61 & 27.71 & 28.72 & - & - & - & - & - & -   \\
\devc{}* (mean)&  - & - & - & 28.17 & 27.17 & 28.39 & 28.24 & 27.30 & 28.44 \\
\devc{}* (linear)&  - & - & - & 28.35 & 27.39 & 28.53 & 28.54 & 27.68 & 28.69  \\
Ours w/o tracking& 28.61 & 27.71 & 28.72 & 28.37& 27.50 & 28.54 & 28.58 & 27.76 & 28.73  \\
Ours (inst.)& 28.59 & 27.75 & 28.70 & 28.36 & 27.55 & 28.52 & 28.58 & 27.82 & 28.72 \\
Ours (dense)& {\bf 28.74} & 29.12 & 27.79 & {\bf 29.27} & 27.59 & {\bf 28.94} & 29.02 & 27.85 & {\bf 29.19} \\
Ours (inst.+dense)& {\bf 29.13} & {\bf 27.94} & {\bf 29.27} & {\bf 28.74} & {\bf 27.69} & 28.93 & {\bf 29.03} & {\bf 27.97} & {\bf 29.19} \\

\bottomrule
\end{tabular}
\caption{PSNR. Higher is better. The first block of three columns shows the results with single reference frame. The remaining two blocks show the results with multi-reference setting. Full, Inner, and Outer refer to the region with respect to instance masks used for evaluation.} %We use instance mask annotation of DAVIS for instance-level evaluation (Inner).}
\label{table:psnr}
\end{center}
\vspace{-6mm}
\end{table*}
%%%%%%%%%%

Fig.~\ref{fig:algo_dense} visualizes how we propagate masks using dense tracking. To calculate the mask $M^D_{r}\in \mathbb{R}^{HW \times HW}$, which represents candidate positions at a reference frame $r$ to warp to the $i$-th position of $x_t$, we repeat propagating a mask from a target $x_t$ to a reference $y_r$.
Let $t' \in \{t, t+1, \dots, t_r\}$ be a set of frame indices from $x_t$ to $y_r$, where $t_r$ represents a frame index of $y_r$, and $D^{t'}(i_0) \in \{0, 1\}^{HW}$ be propagated candidate positions at $t'$-th frame from the $i_0$-th position of $x_t$. We define the initial mask at $t'=t$ as it has value $1$ only at $i=i_0$ position otherwise 0:
\begin{equation}
D^{t}_{i}(i_0) = 
\begin{cases}  
1& \text{if }\ i = i_0 \\ 
0& \text{otherwise}  
\end{cases} \\
\end{equation}
Assuming that the moving destination of the pixel exists in the nearest neighbors, we propagate candidate positions from the $i_0$-th position of $x_t$ to a reference frame $y_r$ by following equations:
\begin{align}
    \label{eqn:binarize}
    &D^{t'}_j(i_0) = \mathrm{binarize}\biggl(\sum_{i}\frac{A^D_{i,j}(x_{t'-1}, x_{t'})}{\sum_{j'}A^D_{i,j'}(x_{t'-1}, x_{t'})}\ D^{t'-1}_{i}(i_0)\biggr) \\
    &A^D_{i,j}(x_{t'-1}, x_{t'}) = 
    \begin{cases}
        A_{i,j}(x_{t'-1}, x_{t'}) &\text{if }\  j \in \mathcal{N}_i^R\\
        0   &\text{otherwise}
    \end{cases}
\end{align}
where $A^D \in [0, 1]^{HW \times HW}$ is a spatially restricted affinity, and $\mathcal{N}_i^R$ is a set of neighboring pixels residing within the window whose size is $2R \times 2R$ centered at $i$-th position. Here, $R$ is a hyperparameter to limit neighbors. Propagating masks from $t$-th frame to $t_r$-th frame, we calculate a dense tracking mask $M^D_r \in \{0, 1\}^{HW \times HW}$ for a reference frame $y_r$ as follows:
\begin{equation}
M^D_{r, i, j} =  
\begin{cases}  
1& \text{if } D^{t_r}_{j}(i) = 1 \\ 
0& \text{otherwise}  
\end{cases} \\
\end{equation}

As described in equation \ref{eqn:binarize}, in order to generate many candidates for tracking destinations, we convert pixels, whose similarity score exceeds a threshold value, to binary masks as candidates for movement. By creating a one-to-many relationship between the starting point and the destinations of tracking, it encourages the generation of multiple mask candidates, even when starting at only one point in the target. Furthermore, by setting binary masks with threshold, it is possible to prevent the repeated multiplication of the probability from reaching zero, which frequently happens without performing threshold processing.

By repeating the above update from target to reference, for each pixel on the target frame, a mask representing the destination on reference is obtained. As a result, it is possible to restrict regions as a candidate for soft warp of a color.

%%%%% Figure %%%%%
\begin{figure*}
\begin{center}
\includegraphics[width=\linewidth]{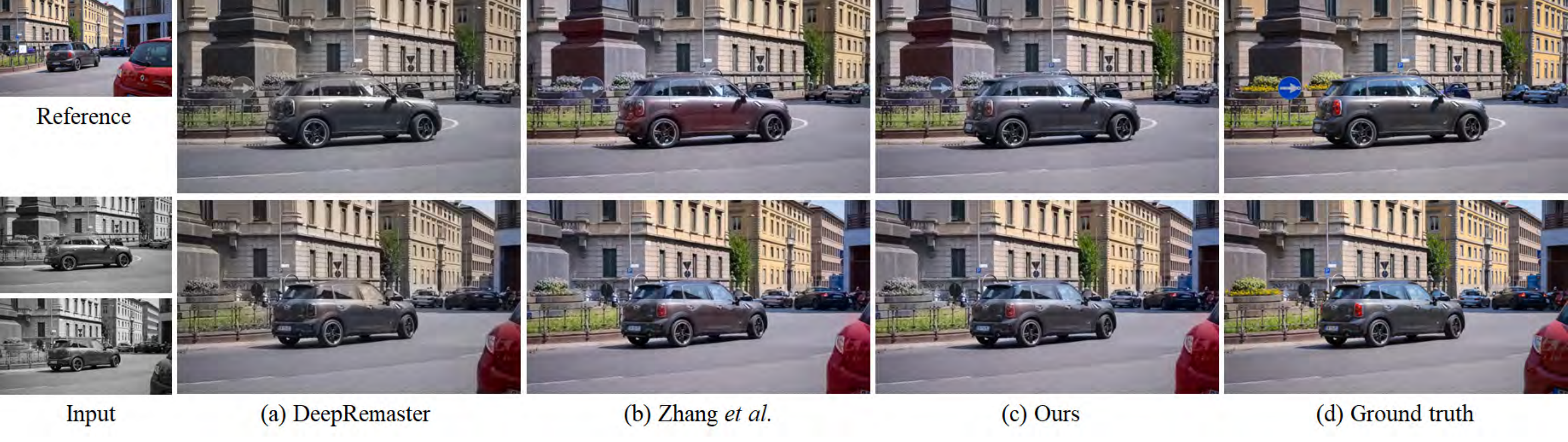}
\end{center}
   \caption{Qualitative comparisons with state-of-the-arts with single reference. When there are two car on the reference frame, the body of the car on (b) \devc{} contains red. On the other hand, (c) ours is not affected by the red car as we identify the car instances, and the color is clearer than (a) DeepRemaster\cite{deepremaster}.}
\label{fig:quality_single}
%\end{figure*}
%%%%%
%%%%% Figure %%%%%
%\begin{figure*}
\begin{center}
\includegraphics[width=\linewidth]{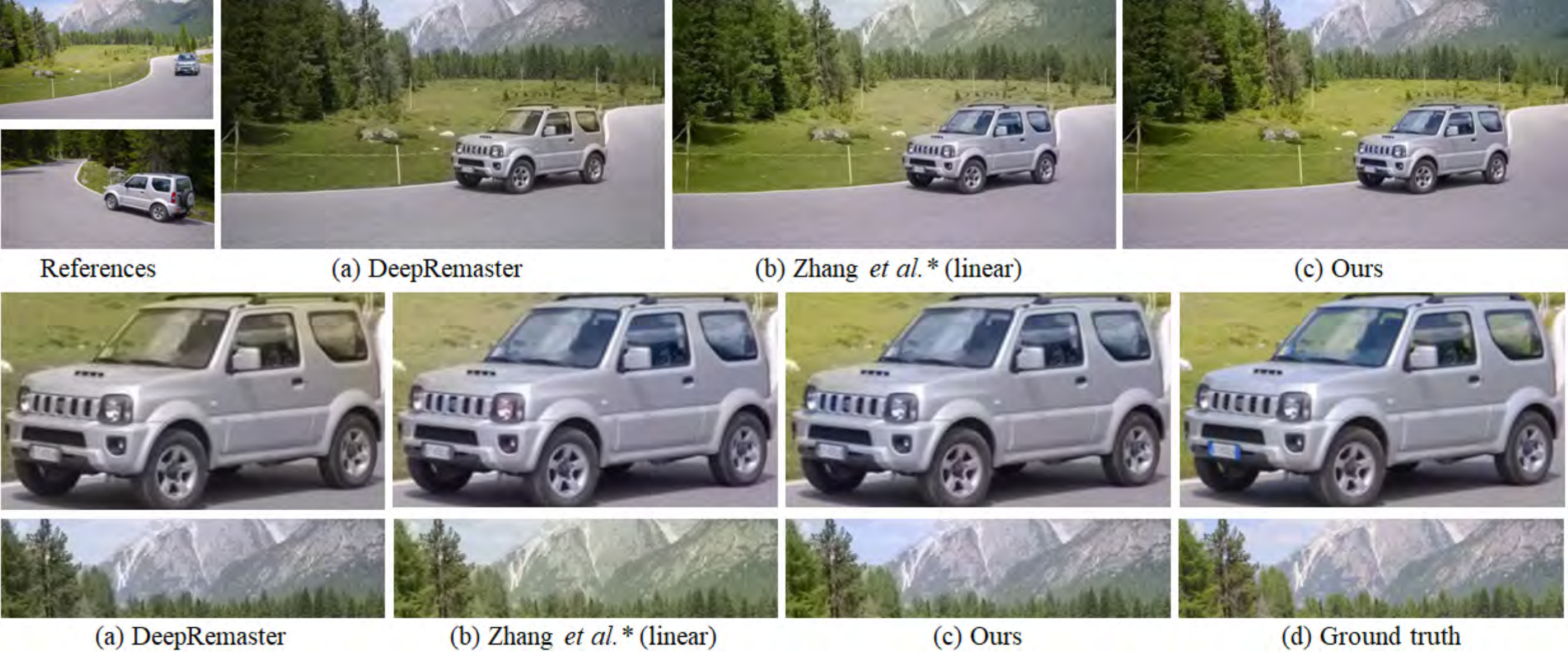}
\end{center}
   \caption{Qualitative comparisons with state-of-the-arts with multiple reference. (a) DeepRemaster\cite{deepremaster}'s car is green and the sky and the trees are faded. On (b) \devc{}* (linear), the car headlight seems red, and the sky and the mountains are green.}
\label{fig:quality_multi}
\end{figure*}
%%%%%

%%%%% Figure %%%%%
\begin{figure}
\begin{center}
\includegraphics[width=\linewidth]{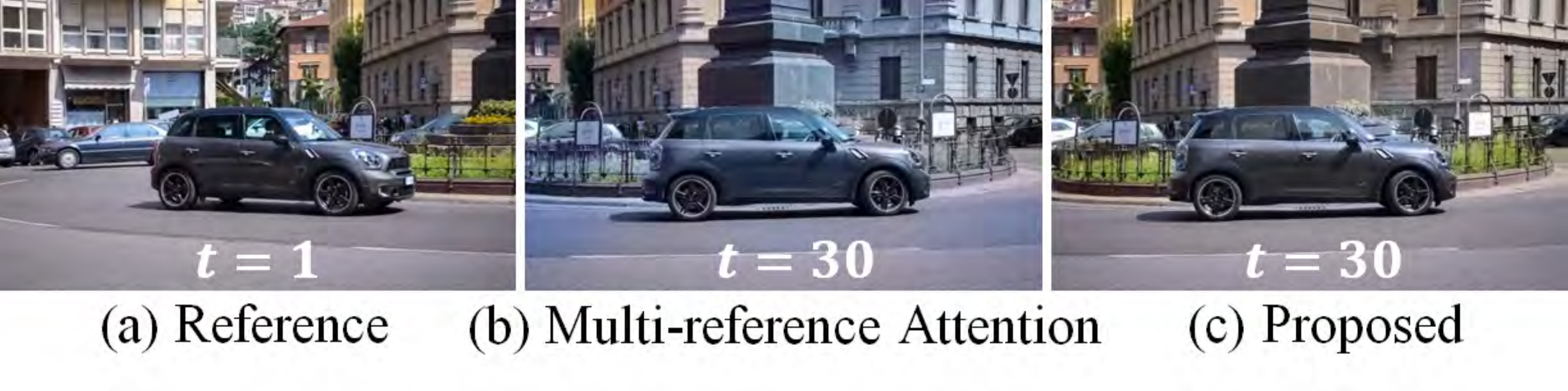}
\end{center}
   \caption{Comparison of color propagation and mask propagation. (b) shows the result of Fig. \ref{fig:approach}(c), and color propagation errors accumulate visibly . Conversely, plausible color can be warped from the reference frame even with an error in the mask propagation (c).}
\label{fig:temporal_corr}
\end{figure}
%%%%%

%%%%% Figure %%%%%
\begin{figure}
\begin{center}
\includegraphics[width=\linewidth]{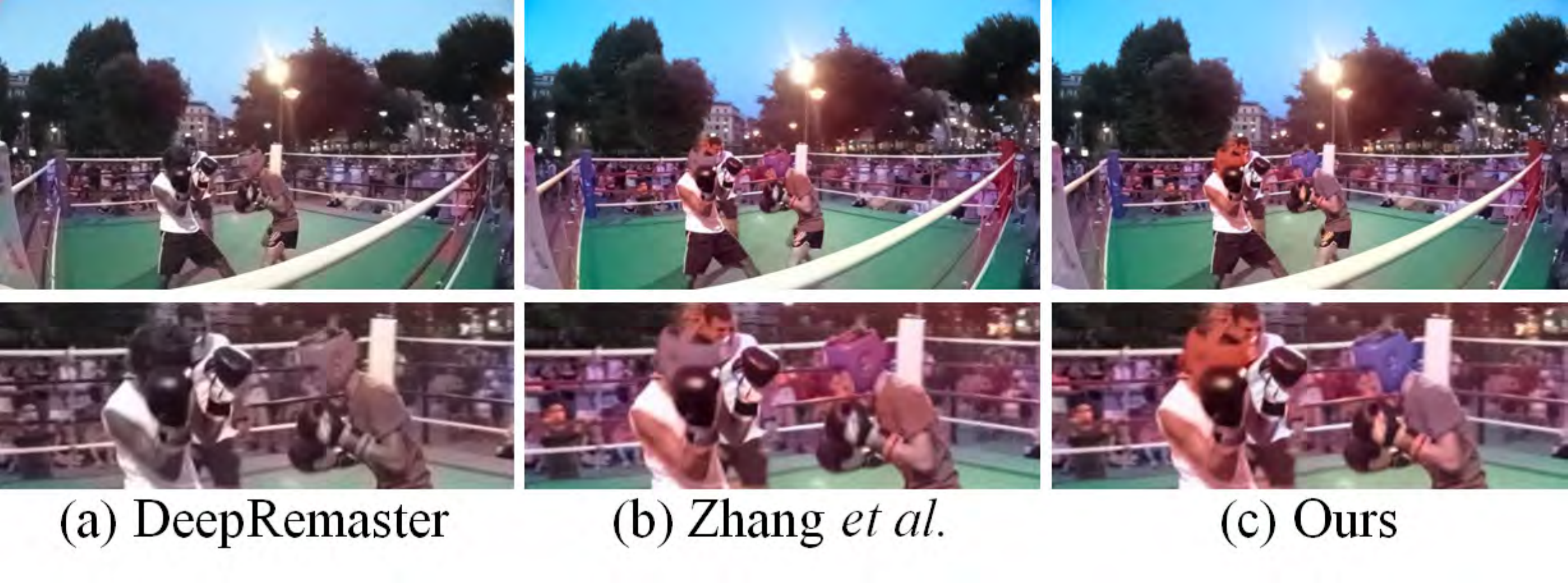}
\end{center}
   \caption{A scene where instance tracking is effective.}
\label{fig:comp_inst}
\end{figure}
%%%%%

% ---------------------------------------------
\section{Experiments}

% -----
\subsection{Experimental Setting}
{\bf Datasets.} 
We mainly report our results on DAVIS-2017 \cite{davis2017}. We refer the readers to Appendix for results on other datasets, as they generally exhibit similar trends. We test video colorization on 90 videos from the training and validation splits. We resize videos to $384 \times 216$, and set window size $R = 9$ and $\text{threshold} = 0.2$ in binarize function. Note that none of the data from DAVIS-2017 are used for training. In DAVIS-2017, ground truth instance segmentations are annotated at pixel level in each frame.

{\bf Evaluation Metrics.}
As with the existing quantitative evaluation protocol for colorization methods, we report the peak signal-to-noise ratio (PSNR) to compare with other methods. It is possible to evaluate whether the color can be propagated faithfully to the reference color since, by providing ground truth colors as a reference frame, it is expected that the ground truth color is propagated to the following frames.
Note that video with frequent dynamic changes cannot fully reproduce the ground truth frame because not all colors can be specified in a limited number of reference frames.

{\bf Network Training.}
Since our network relies on \cite{devc} as the backbone, we evaluate it using its distributed parameters to make fair comparisons. The feature extraction network for calculating the correspondence and the refinement network are trained end-to-end using multiple effective losses \cite{devc}.
As described in Sec \ref{method:overview}, no additional training is required to perform our dense tracking.

% -----
\subsection{Comparisons with the state-of-the-art}

{\bf Quantitative Comparisons.}
We report the results for PSNR on DAVIS in Table \ref{table:psnr}. We report experiments with three types of reference frames: 1) single reference of 10th frame from the beginning, 2) multiple references given at the beginning (1st) and the end ($\textit{N}$-th) of a sequence, and 3) multiple references at the 10th frame from the beginning and 10th from the end ($\textit{N}$-10th). Note that the position of reference frame can be arbitrary, as shown in Appendix.
As well as PSNR for the entire image, we also report the evaluations inside and outside the ground truth instance masks on DAVIS.
Note that while \cite{deepremaster} consists of two networks, a restoration network and a colorization network, we use only the colorization network.
Since \devc{} proposes a workflow for a single reference, in order to compare the multiple reference results with \devc{}, we create the outputs at each reference, and assign them weights either by mean or linearly as the distance between the target and the reference frame. We refer to them as \devc{}* (mean) and \devc{}* (linear), respectively.

We report four variations of our method.
Ours (w/o tracking) is an extension of \devc{} to a mechanism for multiple references.
When only instance tracking is used (Ours (inst.)), an improvement in the inner of the instance mask is observed. Note that there is a change in the score outside the mask, because the region detected by the instance detector is different from the instance region of the ground truth.
Ours (dense) uses dense tracking to improve scores across images. Ours (inst. + dense), which further improved on instance tracking, achieves the highest performance in most cases.
These results demonstrate that using temporal correspondence and restricting the region to reflect the reference frame improve performance.

%%%% Figure %%%%%
\begin{figure*}
\begin{center}
\includegraphics[width=\linewidth]{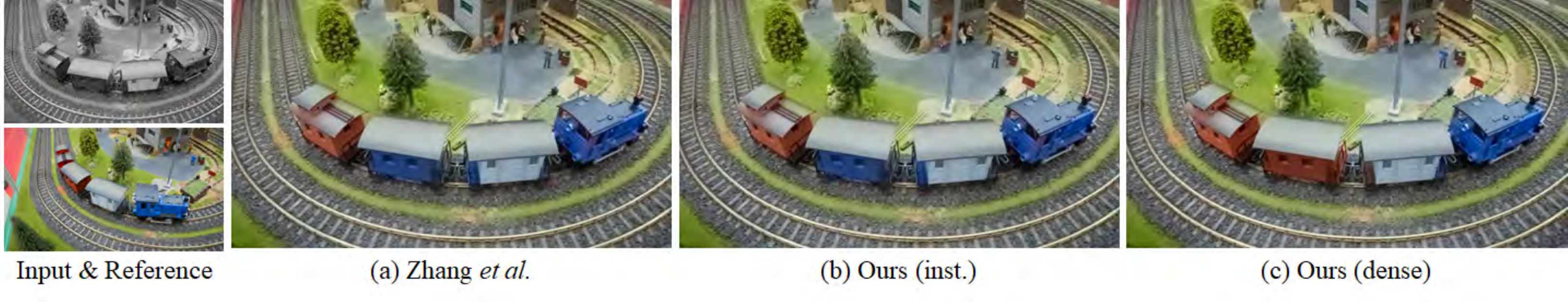}
\end{center}
   \vspace{-5pt}
   \caption{Comparison of dense tracking with other methods. (a) \devc{} using non-local attention struggles with long time or changing objects. (b) Instance detection is not successful in this case, and the result also affects colorization. On the other hand, by tracking on a vehicle-by-vehicle basis, (c) ours (dense) propagates correct colors. }
\label{fig:comp_dense}
\end{figure*}
%%%%%

%%%% Figure %%%%%
\begin{figure}
\begin{center}
\includegraphics[width=\linewidth]{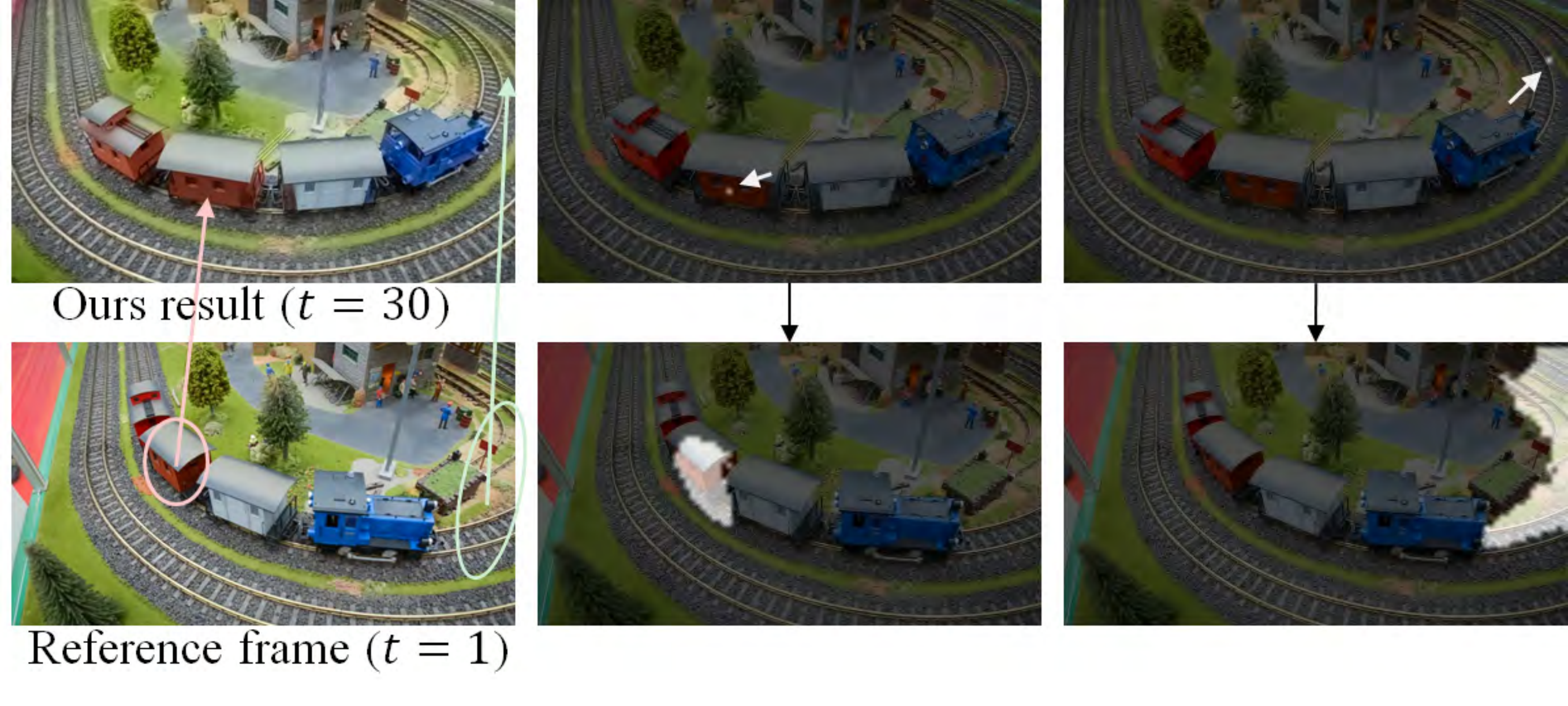}
\end{center}
   \caption{Visualization of dense tracking. We use dense tracking to obtain masks on the reference frame corresponding to each pixel on the target frame. We determine a color of the target pixel by further weighting from masked regions on the reference image.}
\label{fig:vis_dense}
\end{figure}
%%%%%

{\bf Qualitative Comparisons.}
Fig. \ref{fig:quality_single} and Fig. \ref{fig:quality_multi} show comparisons of the results of the proposed method and the state-of-the-art methods. Fig. \ref{fig:quality_single} shows the results when there are the same semantic objects (car) in a reference frame. 
\devc{} warps colors from the two cars, and as a result, the car in the target frame incorrectly contains red. The stone tower on the left-hand side of the image also incorrectly contains red. In contrast, our method prevents color transfer from the red car by considering the temporal relationship between frames. In addition, the color is clearer than that of DeepRemaster \cite{deepremaster}.
As shown in Fig. \ref{fig:quality_multi}, our method achieves more accurate colorization than other methods even when there are multiple reference images. Our proposed framework with spatiotemporal correspondence improves colorization performance at a finer level than instances, such as the color of the pixel of the sky and mountains.

% -----
\subsection{Analysis}

{\bf Spatiotemporal Correspondence.}
Our approach expresses temporal correspondence as mask propagation. Another approach (Fig.~\ref{fig:approach}(c)) that warps colors from both the reference frame and previous frame tends mostly to use the color of the previous frame because the previous frame is more similar to current target frame than the reference frame. In this method, therefore, it is difficult to colorize a long video due to accumulation of errors, as shown in Fig. \ref{fig:temporal_corr}. On the other hand, our method results in high fidelity and can colorize a long video, by warping the color of the reference frame, while considering the relation between the target frames.

{\bf Instance Tracking.}
Fig. \ref{fig:comp_inst} shows the effect of instance tracking. Our model takes advantage of the highly accurate segmentation of off-the-shelf instance detection. Our model can also prevent color leakage between instances even in long videos. Table \ref{table:psnr} shows that the performance can be improved within the ground truth instance mask.

{\bf Dense Tracking.}
Fig. \ref{fig:comp_dense} compares dense tacking with other models. Dense tracking tracks every pixel on the target frame. Fig. \ref{fig:vis_dense} visualizes dense tracking. As shown in the figures, tracking can be performed regardless of the class or size of the object.
The color is determined by calculating the similarity only in the masked region propagated from the target frame to the reference frame.
Therefore, while the tracking accuracy is not as high as with instance tracking, the influence of the error can be reduced at the time of warping the color.

% ---------------------------------------------
%\section{Discussion}
% {\bf Legacy video.} Fig. X shows the result for legacy video by our method and the manual colorization by a professional creator. The results are 30th frame later from the reference frame. The results present that our method is effective for large cost reduction. 

{\bf Limitations.} 
In our method, thresholds for window to limit neighbors are determined manually for dense tracking. If window size is too large, they can span different instances. On the other hand, if the window size is too small, it is difficult to track an object having a large amount of movement. Therefore, it is necessary to determine a threshold adaptively according to the content of the video. For our future work, we are interested in automating the setting of manual thresholds such as window size.
% We may add / replace future works with the following:
% * Finetuning network with spatio-temporal correspondence mechanism
% * make hyper parameters of instance tracking & dense tracking (window size, threshold etc.) differentiable 

% ---------------------------------------------
\section{Conclusions}
We proposed a novel framework for reference-based video colorization.
Our method uses temporal correspondence between target frames to reflect the reference color more faithfully.
We also proposed a dense tracking method in addition to the introduction of instance tracking as a method to compute temporal correspondence.
Experimental results demonstrate that our model outperforms state-of-the-art models both qualitatively and qualitatively. While it is still challenging to track across different scenes with different camera shots, it may be handled efficiently by detecting scene switches, which remains as our future work.
%We hope that our proposed framework accelerates the integration with other cutting-edge computer vision technologies for practical video colorization.

%For future work, it is still difficult to track across different scenes with different camera shots, but it can be extended more practically by detecting scene switches. Some of the existing computer vision technologies, such as re-ID, can be easily incorporated into our framework and enhance the performance.

\subsection*{Acknowledgement}
We thank Masato Ishii for helpful discussions and comments. We also thank Sony Pictures Entertainment Technology Development Group for providing data and feedback.

%%%%%%%%% END OF BODY TEXT
{\small
\bibliographystyle{ieee_fullname}
\bibliography{egbib}
}

\clearpage
\appendix
%\appendixpage
%%%% BODY TEXT

\section{Effect of Reference Frame's Position}
Table \ref{table:psnr_davis_additional} shows the results of PSNR on DAVIS-2017 dataset \cite{davis2017}, using different reference frame positions from the one reported in the main paper.
We only use 30 frames per a video on the single reference setting, since colors that can be specified by a single reference may not include the ground truth colors of distant frames, although more frames can be colored.
% explain 1st and Nth frame? and Full/Inner/Outer

As shown in the table, regardless of the reference frame's position within the video, our model consistently outperforms previous models, demonstrating that it is resilient to the change of reference frame's position.

\section{PSNR on Videvo Dataset}
Table \ref{table:psnr_videvo} shows that our method can better-reflect the colors more from ground truth than the previous methods in Videvo dataset.
Videvo dataset includes 60 videos (30 frames/video) that we collected manually from Videvo website\footnote{https://www.videvo.net} such that it includes various scenes. Unlike DAVIS-2017, there is no ground truth segmentation mask, so we only evaluate the entire frame (Full) for each frame.
Our observation is that instance detection by MaskRCNN \cite{maskrcnn} is not successful for Videvo dataset, due to a large number of instances moving or substantial blurs in the background. We suspect that this is why there is no difference in scores between Ours(dense) and Ours(inst.+dense). Also note that, in consistency with Table \ref{table:psnr_davis_additional}, our model outperforms previous models regardless of the reference frame's position.

\section{Other Evaluation Metric}

We report the results with another evaluation metric for reference-based video colorization.
The metric we introduce is the percentage of pixels whose error magnitude exceeds a threshold. This is similar to how the KITTI benchmark \cite{Menze2015CVPR} evaluates the correctness of estimated disparity or flow end-point error (often called \% Outlier). In particular, this metric is designed to measure the user controllabilty. For example, given a case where the creator needs to partially change the color of the area after applying the colorization model, this metric provides a clue as to how much revision will have to be made by the user.

\begin{figure*}
\begin{center}
\includegraphics[width=0.6\linewidth]{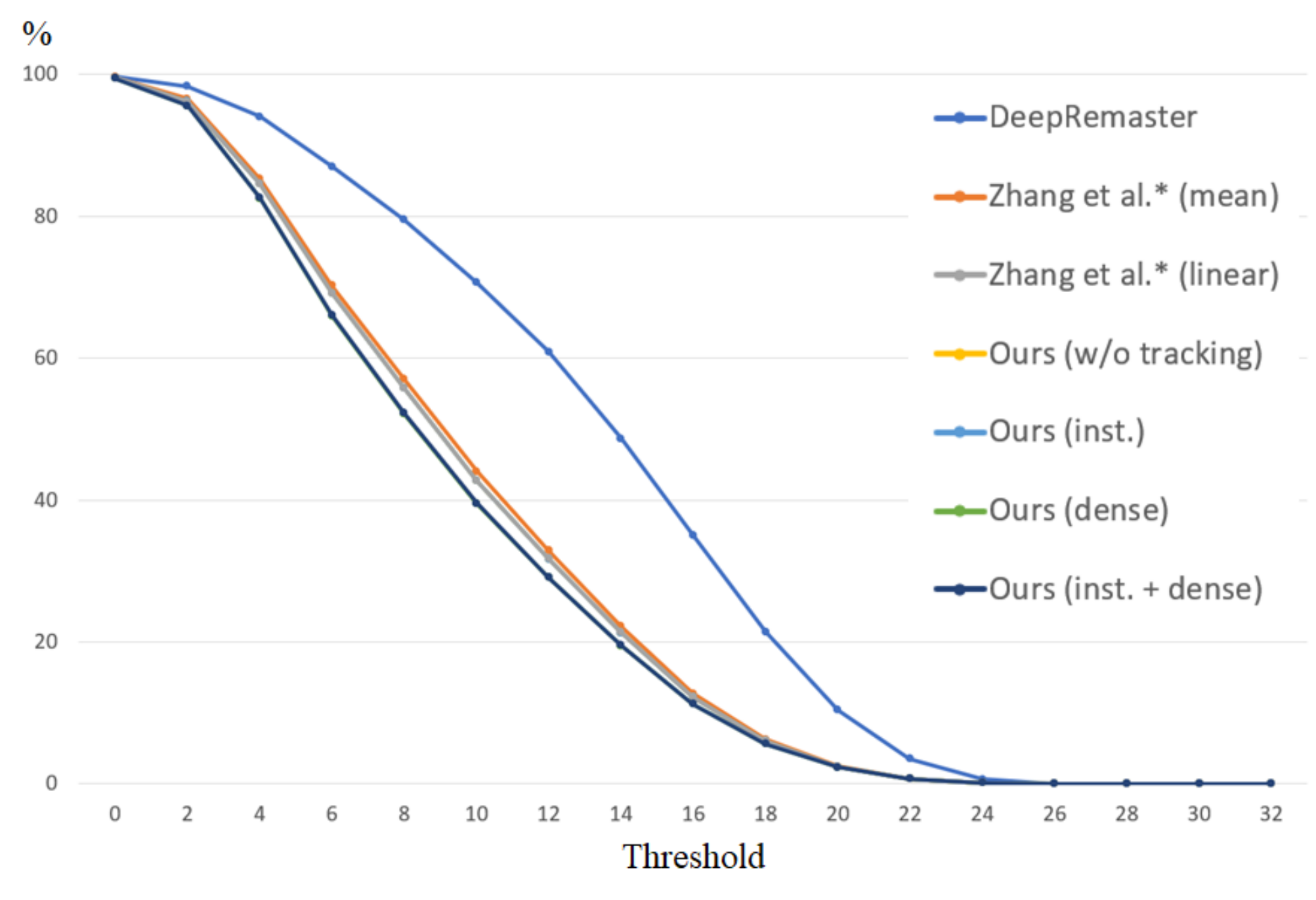}
\end{center}
   \caption{This is the percentage of pixels that is over a threshold of Euclidean distance from ground truth in RGB space.}
\label{fig:supp_outlier}
\end{figure*}

Fig.~\ref{fig:supp_outlier} shows the relationship between the threshold and the percentage of pixels where the error exceeds the threshold. The percentages are averaged across the entire DAVIS-2017 \cite{davis2017} dataset.
Ours has consistently fewer errors than other methods. This demonstrates that our method is more similar to ground truth color and realizes better user-controllability and lower costs.
For example, in \devc{}* (linear) vs Ours (inst.+dense), there is a 3.5\% difference,  when $\textit{Threshold} = 8$.
This implies that at 4K (8.3 million pixels), there is a difference of about 290,000 pixels. In other words, our model leads to reduction of user re-drawing effort of about 540$\times$540 pixels.

\begin{figure*}
\begin{center}
\includegraphics[width=\linewidth]{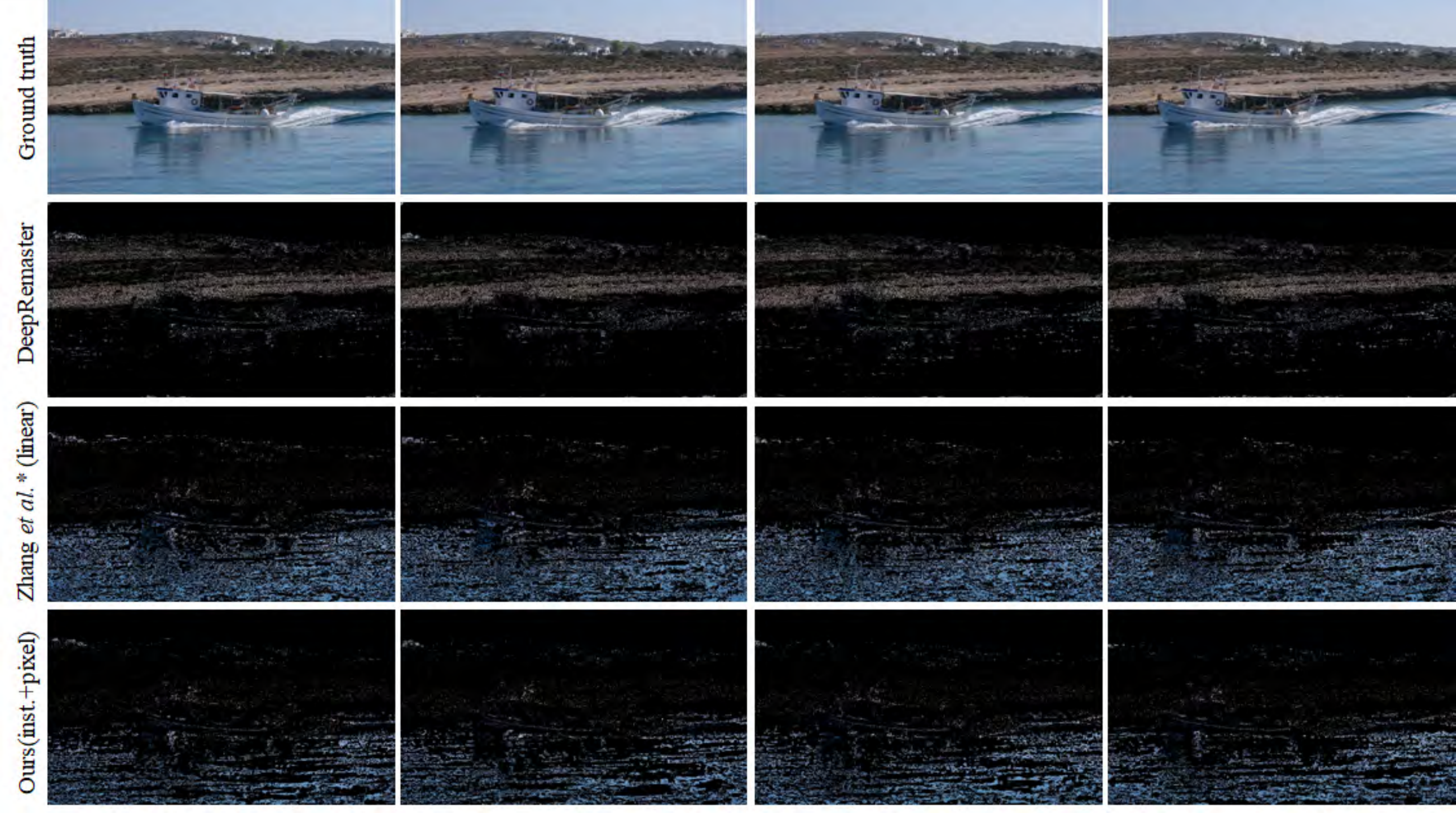}
\end{center}
   \caption{The visualization of the pixels that is over a threshold of Euclidean distance from ground truth in RGB space. We set the $\textit{threshold} = 16$. Sparser visualization indicates better performance.}
\label{fig:supp_vis_outlier_1}
\end{figure*}

\begin{figure*}
\begin{center}
\includegraphics[width=\linewidth]{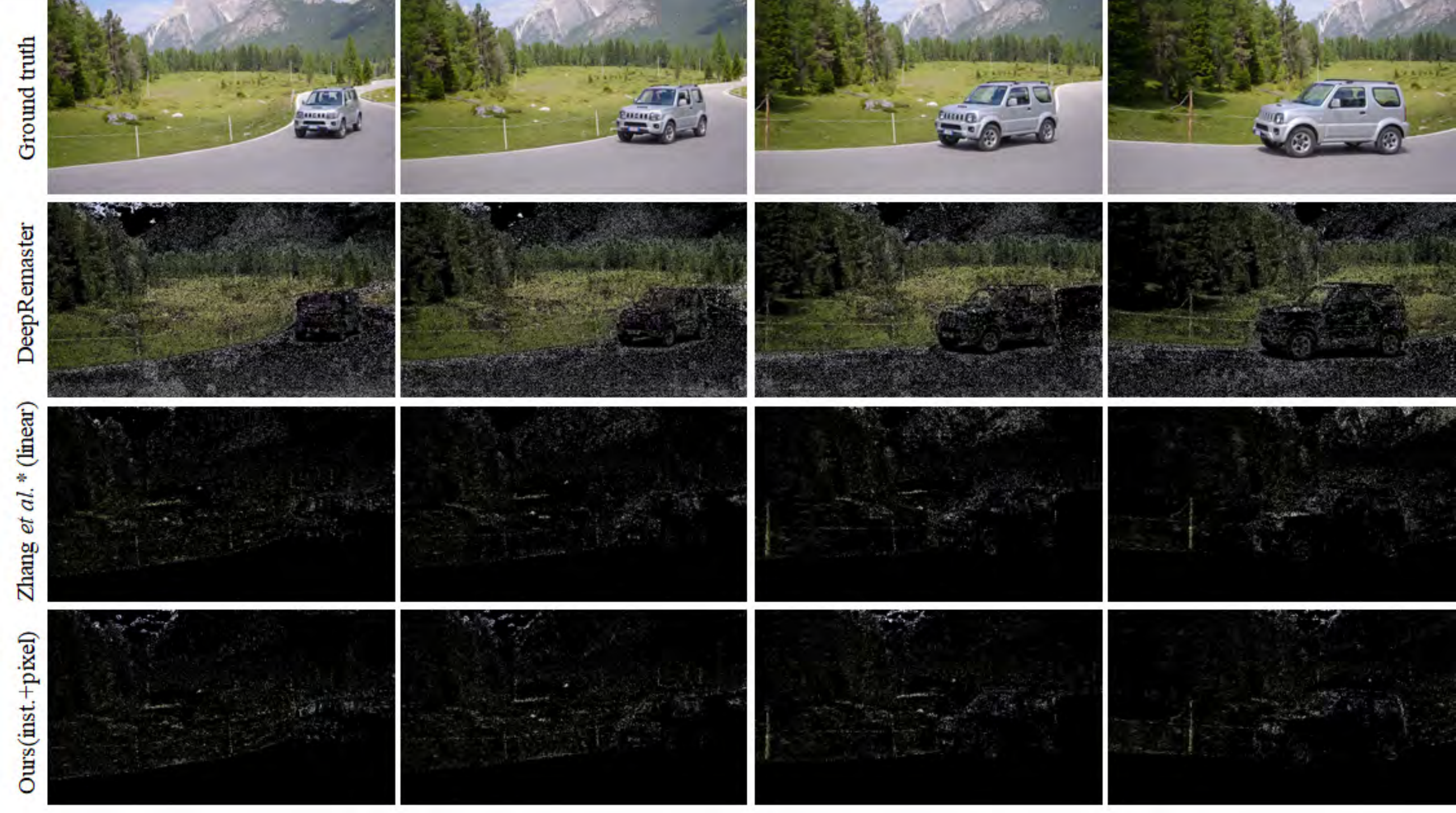}
\end{center}
   \caption{The visualization of the pixels that is over a threshold of Euclidean distance from ground truth in RGB space. We set the $\textit{threshold} = 16$. Sparser visualization indicates better performance.}
\label{fig:supp_vis_outlier_2}
\end{figure*}

Table \ref{table:outlier_davis} and Table \ref{table:outlier_videvo} are the results on DAVIS and Videvo respectively, when we set $\textit{Threshold} = 16$. Our method performs better than the previous methods, and in particular, we can see that dense tracking is critical for enhancing the performance for the entire image. Fig. \ref{fig:supp_vis_outlier_1} and  Fig. \ref{fig:supp_vis_outlier_2} show visualization of pixels over a threshold of Euclidean distance from ground truth in RGB space, comparing our model with previous models.

%%%%%%%%%% Table
\begin{table*}
\begin{center}
\begin{tabular}{lcccccc}
\toprule
&\multicolumn{6}{c}{Single Reference}\\
\midrule
&\multicolumn{3}{c}{1st frame}&
\multicolumn{3}{c}{Nth frame}  \\
Method & Full & Inner & Outer & Full & Inner & Outer \\
\midrule
DeepRemaster \cite{deepremaster}& 25.34 & 24.23 & 25.51 & 25.31 & 24.09 & 25.49 \\
\devc{} & 28.32 & 27.48 & 28.41 & 28.06 & 27.21 & 28.15 \\
Ours (inst.)& 28.31 & 27.52 & 28.40 & 28.00 & 27.21 & 28.15 \\
Ours (dense)& 28.75 & 27.56 & 28.87 & {\bf 28.51} & 27.36 & {\bf 28.64} \\
Ours (inst.+dense)& {\bf 28.76} & {\bf 27.68} & {\bf 28.88} & 28.45 & {\bf 27.40} & 28.63 \\

\bottomrule
\end{tabular}
\caption{PSNR on DAVIS. Higher is better. We report experiments with two types of reference frames: 1) single reference given at the beginning (1st), and 2) the end ($\textit{N}$-th) of a sequence.
Full, Inner, and Outer refer to the region with respect to ground truth instance masks used for evaluation.} % The first block of three columns shows the results with single reference frame. The remaining two blocks show the results with multi-reference setting. Full, Inner, and Outer refer to the region with respect to instance masks used for evaluation.We use instance mask annotation of DAVIS for instance-level evaluation (Inner).}
\label{table:psnr_davis_additional}
\end{center}
%\vspace{-6mm}
\end{table*}
%%%%%%%%%%

%%%%%%%%%% Table
\begin{table*}
\begin{center}
\begin{tabular}{lcccc}
\toprule
&\multicolumn{3}{c}{Single Reference} & Multiple References\\
\midrule
&\multicolumn{1}{c}{1st frame}&
\multicolumn{1}{c}{10th frame}&
\multicolumn{1}{c}{Nth frame}&
1st \& Nth frames\\
Method & Full  & Full &  Full & Full \\
\midrule
DeepRemaster \cite{deepremaster}& 23.88 & 24.08 & 23.85 & 24.22 \\
\devc{} & 26.98 & 27.33 &  26.77 & - \\
\devc{}* (mean) & - & - & - & 27.41 \\
\devc{}* (linear) & - & - & - & 27.63 \\
Ours (w/o tracking) & 26.98 &  27.33 &  26.77  & 27.64 \\
Ours (inst.) & 26.97 & 27.32 & 26.70 & 27.65 \\
Ours (dense) & {\bf 27.35} & {\bf 27.76} & {\bf 27.11} & 28.09 \\
Ours (inst.+dense)& {\bf 27.35} & {\bf 27.76} & 27.04 & {\bf 28.10} \\

\bottomrule
\end{tabular}
\caption{PSNR on Videvo. Higher is better.} % The first block of three columns shows the results with single reference frame. The remaining two blocks show the results with multi-reference setting. Full, Inner, and Outer refer to the region with respect to instance masks used for evaluation.We use instance mask annotation of DAVIS for instance-level evaluation (Inner).}
\label{table:psnr_videvo}
\end{center}
%\vspace{-6mm}
\end{table*}
%%%%%%%%%%

%%%%% Figure %%%%%
%%%%%

%%%%%%%%%% Table
\begin{table*}
\begin{center}
\begin{tabular}{lccccc}
\toprule
&\multicolumn{3}{c}{Single Reference} & \multicolumn{2}{c}{Multiple References}\\
\midrule
Method &\multicolumn{1}{c}{1st} & \multicolumn{1}{c}{10th}& \multicolumn{1}{c}{Nth} & 1st \& Nth & 10th \& N-10th\\
\midrule
DeepRemaster \cite{deepremaster}& 36.0 & 35.7 & 35.9 & 35.3 & 35.1 \\
\devc{} & 12.7 & 12.2 & 13.0 & - & - \\
\devc{}* (mean) & - & - & - & 12.8 & 12.7 \\
\devc{}* (linear) & - & - & - & 12.5 & 12.3 \\
Ours (w/o tracking) & 12.7 & 12.2 & 13.0 & 12.5 & 12.2 \\
Ours (inst.) & 12.8 & 12.2 & 13.0 & 12.5 &  12.2 \\
Ours (dense) & {\bf 11.8} & {\bf 11.0} & {\bf 11.9} & {\bf 11.6} & {\bf  11.2} \\
Ours (inst.+dense) & {\bf 11.8} & 11.1 & 12.0 & {\bf 11.6} & {\bf 11.2} \\

\bottomrule
\end{tabular}
\caption{\% Outlier on DAVIS. Lower is better. This is the percentage of pixels that is over a threshold of Euclidean distance from ground truth in RGB space. We set the $\textit{threshold} = 16$.}
\label{table:outlier_davis}
\end{center}
%\vspace{-6mm}
\end{table*}
%%%%%%%%%%

%%%%%%%%%% Table
\begin{table*}
\begin{center}
\begin{tabular}{lcccc}
\toprule
&\multicolumn{3}{c}{Single Reference} & \multicolumn{1}{c}{Multiple References}\\
\midrule
Method &\multicolumn{1}{c}{1st} & \multicolumn{1}{c}{10th}& \multicolumn{1}{c}{Nth} & 1st \& Nth \\
\midrule
DeepRemaster \cite{deepremaster}& 38.4 & 37.8 & 38.3 & 37.9  \\
\devc{} & 14.6 & 13.6 & 14.2 & - \\
\devc{}* (mean) & - & - & - & 13.6 \\
\devc{}* (linear) & - & - & - & 13.1 \\
Ours (w/o tracking) & 14.6 & 13.6 & 14.2 & 12.9 \\
Ours (inst.) & 14.6 & 13.6 & 14.2 & 12.9 \\
Ours (dense) & {\bf 13.8} & {\bf 12.8} & {\bf 13.7} & {\bf 12.1} \\
Ours (inst.+dense) & {\bf 13.8} & {\bf 12.8} & {\bf 13.7} & {\bf 12.1}  \\

\bottomrule
\end{tabular}
\caption{\% Outlier on Videvo. Lower is better. This is the percentage of pixels that is over a threshold of Euclid distance from ground truth in RGB space. We set the $\textit{threshold} = 16$.}
\label{table:outlier_videvo}
\end{center}
%\vspace{-6mm}
\end{table*}
%%%%%%%%%%

%%%%% Figure %%%%%
%%%%%

%%%%% Figure %%%%%
%%%%%

\clearpage

%%%%%%%%% END OF BODY TEX

\end{document}